# High-Quality Prediction Intervals for Deep Learning:
# A Distribution-Free, Ensembled Approach

Tim Pearce [1 2]    Mohamed Zaki [1]    Alexandra Brintrup [1]    Andy Neely [1]

## Abstract

This paper considers the generation of prediction intervals (PIs) by neural networks for quantifying uncertainty in regression tasks. It is axiomatic that high-quality PIs should be as narrow as possible, whilst capturing a specified portion of data. We derive a loss function directly from this axiom that requires no distributional assumption. We show how its form derives from a likelihood principle, that it can be used with gradient descent, and that model uncertainty is accounted for in ensembled form. Benchmark experiments show the method outperforms current state-of-the-art uncertainty quantification methods, reducing average PI width by over 10%.

## 1. Introduction

Deep neural networks (NNs) have achieved impressive performance in a wide variety of tasks in recent years, however, success is generally in terms of aggregated accuracy metrics. For many real-world applications, it is not enough that on average a model performs well, rather the uncertainty of each prediction must also be quantified. This can be particularly important where there is a large downside to an incorrect prediction: Examples can be found in prognostics, manufacturing, finance, weather, traffic and energy networks. There is therefore interest in how NNs can be modified to meet this requirement (Krzywinski & Altman, 2013; Gal, 2016).

In this work the output of prediction intervals (PIs) in regression tasks is considered. Whilst NNs by default output point estimates, PIs directly communicate uncertainty, offering a lower and upper bound for a prediction and assurance that, with some high probability (e.g. 95% or 99%), the realised data point will fall between these bounds. Having this information allows for better-informed decisions.

As an example, a point estimate stating that a machine will fail in 60 days may not be sufficient to schedule a repair, however given a PI of 45-65 days with 99% probability, timing of a repair is easily scheduled.

A diverse set of approaches have been developed to quantify NN uncertainty, ranging from fully Bayesian NNs (BNNs) (MacKay, 1992), to interpreting dropout as performing variational inference (Gal & Ghahramani, 2015). These require either high computational demands or strong assumptions.

In this work we formulate PI output as a constrained optimisation problem. It is self evident that high-quality PIs should be as narrow as possible, whilst capturing some specified proportion of data points (hereafter referred to as the *HQ principle*). Indeed it is through these metrics that PI quality is often assessed (Papadopoulos et al., 2000; Khosravi et al., 2011b; Galván et al., 2017). We show how a loss function can be derived directly from this HQ principle, and used in an ensemble to produce PIs accounting for both model uncertainty and data noise variance. The key advantages of the method are its intuitive objective, low computational demand, robustness to outliers, and lack of distributional assumption.

Notably we build on the work of Khosravi et al. (2011a) who developed the Lower Upper Bound Estimation (LUBE) method, incorporating the HQ principle directly into the NN loss function for the first time. LUBE is gaining popularity in several communities, for example in the forecasting of energy demand and wind speed (section 2). However, we have identified several limitations of its current form.

- **Gradient Descent** - It was stated that the method was incompatible with gradient descent (GD), a belief carried forward, unchallenged, in all subsequent work (section 2). Implementations therefore require non-gradient based methods for training, such as Simulated Annealing (SA) and Particle Swarm Optimisation (PSO). This is inconvenient since GD has become the standard training method for NNs (Goodfellow et al., 2016), used by all modern NN APIs.

- **Loss Form** - Its current form suffers from several problems. The function is at a global minimum when all

---

PIs are reduced to zero. It was also designed through qualitative assessment of the desired behaviour rather than on a statistical basis.

- **Model Uncertainty** - LUBE accounts only for data-noise variance and *not* model uncertainty (section 2.1). This is an oversimplification (Heskes, 1996), implicitly assuming that training data fully populates the input space, which is seldom the case.

In this work we develop a model addressing each of these issues - henceforth referred to as the quality-driven PI method (QD), and QD-Ens when explicitly referring to the ensembled form.

We link early literature on PIs for NNs (Tibshirani, 1996; Heskes, 1996; Papadopoulos et al., 2000; Khosravi et al., 2011a), with recent work on uncertainty in deep learning (Hernández-Lobato & Adams, 2015; Gal & Ghahramani, 2015; Lakshminarayanan et al., 2017) - areas which have remained surprisingly distinct. We achieve this by following the same experimental procedure of recent work, assessing performance across ten benchmark regression datasets. We compare QD's performance with the current best performing model, originally named Deep Ensembles (Lakshminarayanan et al., 2017), here referred to as MVE-Ens. We show that QD outperforms in PI quality metrics, achieving closer to the desired coverage proportion, and reducing average PI width by around 10%.

## 2. Related Work

In this section we consider methods to quantify uncertainty in regression with NNs. Three review papers catalogued early work (Tibshirani, 1996; Papadopoulos et al., 2000; Khosravi et al., 2011b), the latter two specifically considering PIs. Three primary methods were presented:

- The **Delta method** adopts theory for building confidence intervals (CIs) used by general non-linear regression models, estimating *model uncertainty*. It is computationally demanding as it requires use of the Hessian matrix.

- **Mean Variance Estimation** (MVE) (Nix & Weigend, 1994) uses a NN with two output nodes - one representing the mean and the other the variance of a normal distribution, allowing estimation of *data noise variance*. The loss function used is the Negative Log Likelihood (NLL) of the predicted distribution given the data.

- The **Bootstrap** (Heskes, 1996) estimates *model uncertainty*. It trains multiple NNs with different parameter initialisations on different resampled versions of the training dataset. It is easily combined with MVE to estimate total variance.

In addition, BNNs treat model parmeters as distributions rather than point estimates (MacKay, 1992), and hence can predict distributions rather than point estimates. Their drawback is that the computational cost of running MCMC algorithms can be prohibitive. Recent work has focused on addressing this (Graves, 2011; Hernández-Lobato & Adams, 2015; Blundell et al., 2015), notably NNs with dropout may be interpreted as performing variational inference (Gal & Ghahramani, 2015).

Lakshminarayanan et al. (2017) produced a modernisation of Heskes' work (1996), ensembling individual MVE NNs (without resampling the dataset - section 4), and including adversarial training examples. We henceforth refer to this as MVE Ensemble (MVE-Ens). Another MVE extention encourages high uncertainty in data regions not observed by augmenting training data with synthetic 'out-of-distribution' samples of high variance (Malinin et al., 2017).

Many of these modern works complied with an experimental protocol laid out by Hernandez-Lobato & Adams (2015), assessing NLL & RMSE across ten benchmark regression datasets, with MVE-Ens the current best performer. By contrast, the PI literature reports metrics around coverage proportion and PI width.

LUBE (Khosravi et al., 2011a) was developed on the HQ principle. Originally it was proposed with SA as the training method, and much effort has gone toward trialling it with various non-gradient based training methods including Genetic Algorithms (Ak et al., 2013b), Gravitational Search Algorithms (Lian et al., 2016), PSO (Galván et al., 2017; Wang et al., 2017), Extreme Learning Machines (Sun et al., 2017), and Artificial Bee Colony Algorithms (Shen et al., 2018). Multi-objective optimisation has been found useful in considering the tradeoff between PI width and coverage (Galván et al., 2017; Shen et al., 2018).

LUBE has been used in a plethora of application-focused work: Particularly in energy load (Pinson & Kariniotakis, 2013; Quan et al., 2014) and wind speed forecasting (Wang et al., 2017; Ak et al., 2013b), but also prediction of landslide displacement (Lian et al., 2016), gas flow (Sun et al., 2017), solar energy (Galván et al., 2017), condition-based maintenance (Ak et al., 2013a), and others. All work has used LUBE as a single NN, making no attempt to account for model uncertainty (section 2.1).

### 2.1. The Uncertainty Framework

This section describes uncertainty in regression, it is an agglomeration of several prominent works (Tibshirani, 1996; Heskes, 1996; Papadopoulos et al., 2000; Shafer & Vovk, 2008; Mazloumi et al., 2011; Khosravi et al., 2011b; Lakshminarayanan et al., 2017), each of who presented similar concepts but under different guises and terminology. We attempt to reconcile them here.

The philosophy behind regression is that some data generating function, $f(\mathbf{x})$, exists, combined with additive noise, to produce observable target values $y$,

$$y = f(\mathbf{x}) + \epsilon. \qquad (1)$$

The $\epsilon$ component is termed *irreducible noise* or *data noise*. It may exist due to exclusion of (minor) explanatory variables in $\mathbf{x}$, or due to an inherently stochastic process. Some models, for example the Delta method, assume $\epsilon$ is constant across the input space (*homoskedastic*), others allow for it to vary (*heteroskedastic*), for example MVE.

Generally the goal of regression is to produce an estimate $\hat{f}(\mathbf{x})$, which allows prediction of point estimates ($\epsilon$ is assumed to have mean zero). However, when estimating the uncertainty of $y$, additional terms must be estimated. Given that both terms of eq. (1) have associated sources of uncertainty, and assuming they are independent, the total variance of observations is given by,

$$\sigma_y^2 = \sigma_{model}^2 + \sigma_{noise}^2, \qquad (2)$$

with $\sigma_{model}^2$ termed *model uncertainty* or *epistemic uncertainty* - uncertainty in $\hat{f}(\mathbf{x})$ - and $\sigma_{noise}^2$ *irreducible variance*, *data noise variance*, or *aleatoric uncertainty*.

It is worth here distinguishing CIs from PIs. CIs consider the distribution $Pr(f(\mathbf{x})|\hat{f}(\mathbf{x}))$, and hence only require estimation of $\sigma_{model}^2$, whilst PIs consider $Pr(y|\hat{f}(\mathbf{x}))$ and must also consider $\sigma_{noise}^2$. PIs are necessarily wider than CIs.

Model uncertainty can be attributed to several factors.

- **Model misspecification** or **bias** - How closely $\hat{f}(\mathbf{x})$ is able to approximate $f(\mathbf{x})$, assuming ideal parameters and plentiful training data.

- **Training data uncertainty** or **variance** - Training data is a sample from an input distribution. There is uncertainty over how representative the sample is, and how sensitive the model is to other samples.

- **Parameter uncertainty** - Uncertainty exists around the optimum parameters of the model, increasing in regions sparsely represented in the training data.

Different model types have different weightings for each of these factors (*bias-variance trade-off*). Provided the number of hidden neurons is large relative to the complexity of $f(\mathbf{x})$, NNs are considered to have low bias and high variance. Work on uncertainty in NNs therefore generally ignores model misspecification, and only estimates training data uncertainty and parameter uncertainty (Heskes, 1996).

To construct PIs, $\sigma_y^2$ must be estimated at each prediction point. In regions of the input space with more data, $\sigma_{model}^2$ decreases, and $\sigma_{noise}^2$ may become the larger component[1]. In regions of the input space with little data, $\sigma_{model}^2$ grows.

Lakshminarayanan et al. (2017) recognise this in more intuitive terms - that two sources of uncertainty exist.

1. Calibration - Data noise variance in regions which are well represented by the training data.

2. Out-of-distribution - Uniqueness of an input[2] - inputs less similar to training data should lead to less certain estimates.

## 3. A Quality-Driven, Distribution-Free Loss Function

### 3.1. Derivation

We now derive a loss function based on the HQ principle. Let the set of input covariates and target observations be $\mathbf{X}$ and $\mathbf{y}$, for $n$ data points, and with $\mathbf{x}_i \in \mathbb{R}^D$ denoting the $i$th $D$ dimensional input corresponding to $y_i$, for $1 \leq i \leq n$. The predicted lower and upper PI bounds are $\hat{\mathbf{y}}_{\mathbf{L}}, \hat{\mathbf{y}}_{\mathbf{U}}$. A PI should capture some desired proportion of the observations, $(1 - \alpha)$, common choices of $\alpha$ being 0.01 or 0.05,

$$Pr(\hat{y}_{Li} \leq y_i \leq \hat{y}_{Ui}) \geq (1 - \alpha). \qquad (3)$$

A vector, $\mathbf{k}$, of length $n$ represents whether each data point has been captured by the estimated PIs, with each element $k_i \in \{0, 1\}$ given by,

$$k_i = \begin{cases} 1, & \text{if } y_{Li} \leq y_i \leq y_{Ui} \\ 0, & \text{else.} \end{cases} \qquad (4)$$

We define the total number of data points captured as $c$,

$$c := \sum_{i=1}^{n} k_i. \qquad (5)$$

Let Prediction Interval Coverage Probability ($PICP$) and Mean Prediction Interval Width ($MPIW$) be defined as,

$$PICP := \frac{c}{n}, \qquad (6)$$

$$MPIW := \frac{1}{n} \sum_{i=1}^{n} \hat{y}_{Ui} - \hat{y}_{Li}. \qquad (7)$$

---

[1]At the same time, $\sigma_{noise}^2$ may be estimated with more certainty, although uncertainty of this value itself is not generally considered.

[2]Conformal prediction provides a framework to assess this.

According to the HQ principle, PIs should minimise $MPIW$ subject to $PICP \geq (1 - \alpha)$. To minimise $MPIW$, eq. (7) could simply be included in the loss function, however PIs that fail to capture their data point should not be encouraged to shrink further. We therefore introduce *captured* $MPIW$ as the $MPIW$ of *only* those points for which $\hat{y}_L \leq y \leq \hat{y}_L$ holds,

$$MPIW_{capt.} := \frac{1}{c} \sum_{i=1}^{n} (\hat{y}_{Ui} - \hat{y}_{Li}) \cdot k_i. \qquad (8)$$

Regarding $PICP$, we take a likelihood-based approach, seeking NN parameters, $\theta$, that maximise,

$$\mathcal{L}_\theta := \mathcal{L}(\theta | \mathbf{k}, \alpha). \qquad (9)$$

Recognising that each element, $k_i$, is a binary variable taking 1 with probability $(1 - \alpha)$, we represent it as a Bernoulli random variable (one per prediction), $k_i \sim \text{Bernoulli}(1 - \alpha)$. We further assume that each $k_i$ is iid. This independence assumption may not hold for data points clustered close together, however we believe holds sufficiently for a randomly sampled subset of all data points, as used in mini-batches for GD. This iid assumption allows the total number of captured points, $c$, to be represented by a binomial distribution, $c \sim \text{Binomial}(n, (1 - \alpha))$. Substituting in the pmf,

$$\mathcal{L}_\theta = \binom{n}{c} (1 - \alpha)^c \alpha^{n-c}. \qquad (10)$$

The factorials in the binomial coefficient make computation inconvenient. However using the central limit theorem (specifically the de Moivre-Laplace theorem) it can further be approximated by a normal distribution. For large $n$,

$$\text{Binomial}(n, (1 - \alpha)) \approx \mathcal{N}\big(n(1 - \alpha), n\alpha(1 - \alpha)\big) \qquad (11)$$

$$= \frac{1}{\sqrt{2\pi n \alpha (1 - \alpha)}} \exp - \frac{(c - n(1 - \alpha))^2}{2n\alpha(1 - \alpha)}. \qquad (12)$$

We consider this a mild assumption provided a mini-batch size of reasonable number, say $> 50$, is used.

It is common to minimise the NLL rather than maximise the likelihood, this simplifies eq. (12) to,

$$- \log \mathcal{L}_\theta \propto \frac{(n(1 - \alpha) - c)^2}{n\alpha(1 - \alpha)} \qquad (13)$$

$$= \frac{n}{\alpha(1 - \alpha)} ((1 - \alpha) - PICP)^2. \qquad (14)$$

Remembering that a penalty should only occur in the case where $PICP < (1 - \alpha)$ results in a one-sided loss. Combining with eq. (8) and adding a Lagrangian, $\lambda$, controlling the importance of width vs. coverage gives a new loss,

$$Loss_{QD} = $$
$$MPIW_{capt.} + \lambda \frac{n}{\alpha(1 - \alpha)} \max(0, (1 - \alpha) - PICP)^2. \qquad (15)$$

### 3.2. Comparison to LUBE

The derived loss function in eq. (15) may be compared to the LUBE loss (Khosravi et al., 2011a),

$$Loss_{LUBE} = $$
$$\frac{MPIW}{r} \Big( 1 + \exp\big(\lambda \max(0, (1 - \alpha) - PICP)\big) \Big), \qquad (16)$$

where $r = \max(\mathbf{y}) - \min(\mathbf{y})$, is the range of the target variable.

Whilst still recognisable as having the same objective, the differences are significant, and are summarised as follows.

- The inclusion of $n$ intuitively makes sense since a larger sample size provides more confidence in the value of $PICP$, and hence a larger loss should be incurred. Similar arguments follow for $\alpha$. They remove the need to adjust $\lambda$ based on batch size and target coverage.

- A squared term has replaced the exponential. Whilst the RHS for both is minimised when $PICP \geq (1 - \alpha)$, the squared term was derived based on likelihood whilst the exponential term was selected qualitatively.

- $MPIW$ now has an additive rather than multiplicative effect. Multiplying has the attractive property of ensuring both terms are of the same magnitude. However it also means that a global minimum is found when all PIs are of zero width. We found in practise that NNs occasionally did produce this undesirable solution.

- $MPIW$ is no longer normalised by the range of $\mathbf{y}$. Data for a NN should already be normalised during preprocessing. Further normalisation is therefore redundant. It also merely scales the loss by a constant, having no overall effect on the optimisation.

- $MPIW_{capt.}$ is used rather than $MPIW$. As discussed in section 3.1 this avoids the NN benefiting by further reduction of PI widths for missed data.

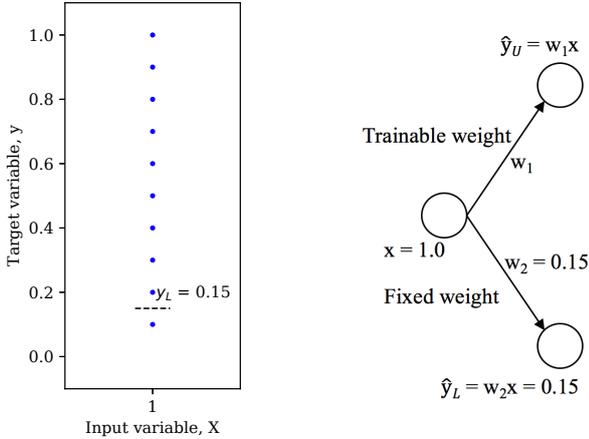

*Figure 1.* Set up for the toy problem: Left is input data (10 data points linearly spaced at fixed input $x = 1.0$), right is the NN used (one trainable weight).

### 3.3. Training QD with Gradient Descent

It was originally believed that the LUBE loss function was, *"nonlinear, complex, and non-differentiable... gradient descent-based algorithms cannot be applied for its minimization"* (Khosravi et al., 2011a). This belief has been carried forward, unchallenged, in all subsequent work - see section 2 for numerous examples. It is inconvenient since GD is the standard method for training NNs, so implementations require extra coding effort.

Regarding the quoted justification, most standard loss functions are nonlinear - e.g. $L_2$ errors - and whilst the LUBE loss function is complex, this does not affect its compatibility with GD[3]. The non-differentiability comment is partially valid. Because the loss function requires the use of step functions, it is not differentiable everywhere. But this is not an unsurmountable problem: ReLUs are a common choice of activation function in modern NNs, despite not being differentiable when the input is exactly zero[4].

#### 3.3.1. GD TOY EXAMPLE

$Loss_{QD}$ can be directly implemented as shown in Algorithm 1 ($Loss_H$), however it fails to converge to a minimum. We demonstrate why this is the case and how it can be remedied through a toy example.

Consider a NN as in figure 1 with one input and two output neurons, linear activations and no bias. For purposes of clarity, one weight is fixed, $w_2 = 0.15$, to create a one dimensional problem with a single trainable weight, $w_1$. Given 10 data points evenly spaced at $x = 1.0$, and $\alpha = 0.2$, the optimal value for $\hat{y}_U$ (and therefore $w_1$) is 0.9, which gives the lowest $MPIW$, subject to $PICP \geq 1 - \alpha = 0.8$.

---

[3]Modern NN APIs generally handle gradient computation automatically, through application of the chain rule to the predefined operations. Provided functions within the API library are used, gradient calculations are automatically handled.

[4]Software implementations return one of the derivatives either side of zero when the input corresponds to the undefined point rather than raising an error (Goodfellow et al., 2016).

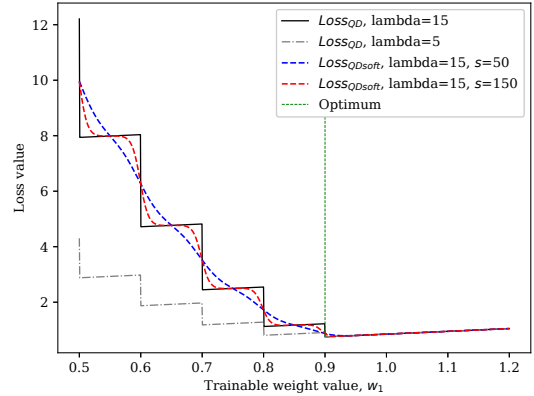

*Figure 2.* Error surface of $Loss_{QD}$, and $Loss_{QD-soft}$ on toy problem, with effect of hyperparameters $\lambda$ and $s$.

$Loss_{QD}$ is plotted in figure 2 (black line). Whilst the global minimum occurs at the desired point, this solution is not found through GD. Given the steepest descent weight update rule with some learning rate $\tau$,

$$w_{1,t+1} = w_{1,t} - \tau \frac{\partial Loss_{QD}}{\partial w_{1,t}}, \qquad (17)$$

the weight shrinks without converging. This is because the gradient, $\frac{\partial Loss}{\partial w_1}$, at any point is positive, except for the discontinuities which are never realised.

To remediate this, we introduce an approximation of the step function. The sigmoid function has been used in the past as a differentiable alternative (Yan et al., 2004). In eq. (4), $\mathbf{k}$, the captured vector was defined. We redefine this as $\mathbf{k}_{hard}$ and introduce a relaxed version as follows,

$$\mathbf{k}_{soft} = \sigma(s(\mathbf{y} - \hat{\mathbf{y}}_L)) \odot \sigma(s(\hat{\mathbf{y}}_U - \mathbf{y})), \qquad (18)$$

where $\sigma$ is the sigmoid function, and $s > 0$ is some softening factor. We further define $PICP_{soft}$ and $Loss_{QD-soft}$ by replacing $\mathbf{k}_{hard}$ with $\mathbf{k}_{soft}$ in equations (6) & (15) respectively - see also $Loss_S$ in Algorithm 1.

Figure 2 shows the result of using $Loss_{QD-soft}$ (red & blue lines). By choosing an appropriate value for $s$, following the steepest gradient *does* lead to a minimum, making GD a

**Algorithm 1** Construction of loss function using basic operations

---

**Input:** Target values, $\mathbf{y}$, predictions of lower and upper bound, $\hat{\mathbf{y}}_L, \hat{\mathbf{y}}_U$, desired coverage, $(1-\alpha)$, and sigmoid softening factor, $s$, $\odot$ denotes the element-wise product.

*# hard uses sign step fn, sign returns -1 if -ve, +1 if +ve*
$\mathbf{k}_{HU} = \max(0, \text{sign}(\hat{\mathbf{y}}_U - \mathbf{y}))$
$\mathbf{k}_{HL} = \max(0, \text{sign}(\mathbf{y} - \hat{\mathbf{y}}_L))$
$\mathbf{k}_H = \mathbf{k}_{HU} \odot \mathbf{k}_{HL}$

*# soft uses sigmoid fn*
$\mathbf{k}_{SU} = \text{sigmoid}((\hat{\mathbf{y}}_U - \mathbf{y}) \cdot s)$
$\mathbf{k}_{SL} = \text{sigmoid}((\mathbf{y} - \hat{\mathbf{y}}_L) \cdot s)$
$\mathbf{k}_S = \mathbf{k}_{SU} \odot \mathbf{k}_{SL}$

$MPIW_c =$
    $\quad \text{reduce\_sum}((\hat{\mathbf{y}}_U - \hat{\mathbf{y}}_L) \odot \mathbf{k}_H)/\text{reduce\_sum}(\mathbf{k}_H)$
$PICP_H = \text{reduce\_mean}(\mathbf{k}_H)$
$PICP_S = \text{reduce\_mean}(\mathbf{k}_S)$
$Loss_H = MPIW_c +$
    $\quad \lambda \cdot \frac{n}{\alpha(1-\alpha)} \cdot \max(0, (1-\alpha) - PICP_H)^2$
$Loss_S = MPIW_c +$
    $\quad \lambda \cdot \frac{n}{\alpha(1-\alpha)} \cdot \max(0, (1-\alpha) - PICP_S)^2$

---

viable method. Setting $s = 160$ worked well in experiments in section 6, requiring no alteration across datasets.

### 3.4. Particle Swarm Optimisation

The original LUBE loss function, eq. (16), has been implemented with various evolutionary training schemes that do not require derivatives of the loss function. In order to test the efficacy of $Loss_{QD-soft}$ with GD, we compared to an evolutionary-based training method (section 5.1). PSO (Kennedy & Eberhart, 1995) was chosen due to use in recent work with LUBE (Galván et al., 2017; Wang et al., 2017). We make the assumption that other evolutionary methods would offer similar performance (Jones, 2005). See Kennedy & Eberhart (2001) for an introduction to PSO.

## 4. Ensembles to Estimate Model Uncertainty

In section 2.1, two components of uncertainty were defined; model uncertainty and data noise variance. It appears that previous work assumed both were accounted for (section 2). In fact, LUBE & QD only estimate data noise variance, and there is a need to consider the uncertainty of these estimates themselves. This becomes particularly important when new data is encountered. Consider a NN trained for the example in figure 1: Despite being capable of estimating the data noise variance at $x = 1.0$, if shown new data at $x = 2.0$ it would predict $\hat{y}_U = 1.8$, with little basis.

Ensembling models provides a conceptually simple way to deal with this. Recall from section 2.1 that three sources of model uncertainty exist, and that the first, model misspecification, is assumed zero for NNs. Parameter uncertainty can be measured by training multiple NNs with different parameter initialisations (*parameter resampling*). Training data uncertainty can be done similarly: Sub-sampling from the training set, and fitting a NN to each subset (*bootstrap resampling*). The resulting ensemble of NNs contains some diversity, and the variance of their predictions can be used as an estimate of model uncertainty.

Recent work reported that parameter resampling offered superior performance to both bootstrap resampling, and a combination of the two (Lee et al., 2015; Lakshminarayanan et al., 2017). No robust justification has been given for this.

Given an ensemble of $m$ NNs trained with $Loss_{QD-soft}$, let $\bar{\mathbf{y}}_U, \bar{\mathbf{y}}_L$ represent the *ensemble's* upper and lower estimate of the PI. We calculate model uncertainty and hence the ensemble's PIs as follows,

$$\bar{y}_{Ui} = \frac{1}{m} \sum_{j=1}^{m} \hat{y}_{Uij}, \qquad (19)$$

$$\hat{\sigma}_{model}^2 = \sigma_{Ui}^2 = \frac{1}{m-1} \sum_{j=1}^{m} (\hat{y}_{Uij} - \bar{y}_{Ui})^2, \qquad (20)$$

$$\tilde{y}_{Ui} = \bar{y}_{Ui} + 1.96\sigma_{Ui}, \qquad (21)$$

where $\hat{y}_{Uij}$ represents the upper bound of the PI for data point $i$, for NN $j$. A similar procedure is followed for $\bar{y}_{Li}$, subtracting rather than adding $1.96\sigma_{Li}$.

## 5. Qualitative Experiments

In this section behaviour of QD is qualitatively assessed on synthetic data. Firstly, the GD method of training explained in section 3.3 is compared to PSO. Next, the advantage of QD over MVE (section 2) in data with non-normal variance is shown. Finally, the effectiveness of the ensembled QD approach at estimating model uncertainty is demonstrated.

The appendix contains experimental details as well as numerical results for this synthetic data, covering a full permutation of loss functions and training methods [LUBE, QD, MVE]x[GD, PSO].

### 5.1. Training method: PSO vs. GD

Comparison of evolutionary methods vs. GD for NN training is its own research topic, and as such analysis here is limited. Preliminary experiments showed that GD performed slightly better than PSO in terms of $PICP$ and $MPIW$, producing smoother, tighter boundaries, both more consistently and with lower computational effort. See figure 3 for a graphical comparison of typical PI boundaries. Data was generated by $y = 0.3 \sin(x) + 0.2\epsilon$, with $\epsilon \sim N(0, x^4)$.

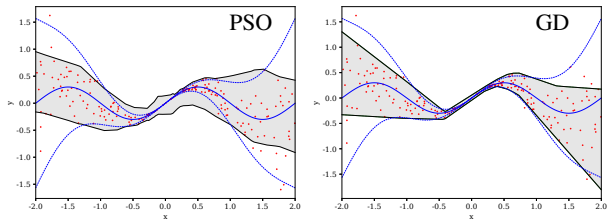

*Figure 3.* Comparison of PI boundaries for GD vs. PSO training methods. Shading is the predicted 95% PI. Ground truth is given by blue lines - the ideal 95% boundaries and mode.

### 5.2. Loss function: QD vs. MVE

Here, the advantage of a distribution-free loss function is demonstrated by comparing MVE, which assumes Gaussian data noise, to QD, which makes no such assumption, on two synthetic datasets. The first was generated as in 5.1 with normal noise, the second with exponential noise, $\epsilon \sim exp(1/x^2)$.

Figure 4 shows, unsurprisingly, that MVE outputs PIs very close to the ideal for normal noise, but struggles with exponential noise. QD approximates both reasonably, though does not learn the boundaries well where data is sparse. Though possible to alter MVE to assume an exponential distribution, this would require significant work. With real data, the distribution would be unknown, and likely irregular, putting QD at an advantage.

### 5.3. Model Uncertainty Estimation: Ensembles

This experiment demonstrated the ability of ensembling to estimate model uncertainty. Data was generated through $y = 0.02x^3 + 0.02\epsilon$, with $\epsilon \sim N(0, 3^2)$. Figure 5 shows ten individual QD PIs as well as the ensembled PIs. The estimated model uncertainty, $\hat{\sigma}^2_{model}$, calculated from eq. (20) is overlaid[5]. Whilst it is difficult to reason about the correctness of the absolute value, its behaviour agrees with the intuition that uncertainty should increase in regions of the input space that are not represented in the training set, here $x \in [-1, 1]$, and $x > 4, x < -4$.

### 6. Benchmarking Experiments

To compare QD to recent work on uncertainty in deep learning, we adopted their shared experimental procedure

---

[5]With uncertainty of the upper and lower bound averaged.

(Hernández-Lobato & Adams, 2015; Gal & Ghahramani, 2015; Lakshminarayanan et al., 2017). Experiments were run across ten open-access datasets. Models were asked to output 95% PIs and used five NNs per ensemble. See appendix for full experimental details. Code is made available online[6].

Previous work reported NLL & RMSE metrics. However, the important metrics for PI quality are $MPIW$ and $PICP$ (section 1). This meant that we had to reimplement a competing method. We chose to compare QD-Ens to MVE-Ens, since it had reported the best NLL results to date (Lakshminarayanan et al., 2017). We did not include LUBE since ensembling and GD had already been justified in section 5.

QD-Ens and MVE-Ens both output fundamentally different things; MVE-Ens a distribution, and QD-Ens upper and lower estimates of the PI. To compute NLL & RMSE for QD-Ens is possible only by imposing a distribution on the PI. This is not particularly fair since the attraction of the method is its lack of distributional assumption. Purely for comparison purposes we did this in the appendix.

A fairer comparison is to convert the MVE-Ens output distributions to PIs, and compute PI quality metrics. This was done by trimming the tails of the MVE-Ens output normal distributions by the appropriate amount, which allowed extraction of $MPIW$, $PICP$, and $Loss_{QD-soft}$. In our experiments we ensured MVE-Ens achieved NLL & RMSE scores at least as good as those reported in the original work, before PI metrics were calculated.

---

[6]https://github.com/TeaPearce

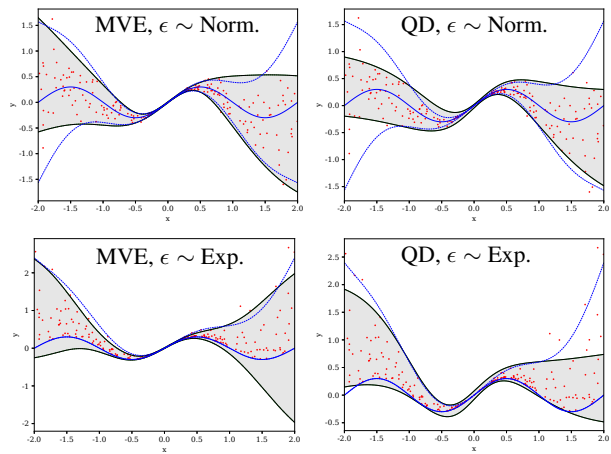

*Figure 4.* Comparison of PI boundaries for two loss functions, QD vs. MVE, given data noise variance drawn from different distributions. Legend as for figure 3.

*Table 1.* PI quality metrics on ten benchmarking regression datasets; mean ± one standard error, best result in bold. Best was assessed according to the following criteria. If $PICP \geq 0.95$ for both, both were best for $PICP$, and best $MPIW$ was given to smallest $MPIW$. If $PICP \geq 0.95$ for neither or for only one, largest $PICP$ was best, and $MPIW$ only assessed if the one with larger $PICP$ also had smallest $MPIW$.

| | Loss$_{QD-soft}$ | | PICP | | MPIW | | |
|---|---|---|---|---|---|---|---|
| | MVE-Ens | QD-Ens | MVE-Ens | QD-Ens | MVE-Ens | QD-Ens | IMPROVEMENT |
| Boston | 1.76 ± 0.28 | **1.33 ± 0.05** | 0.89 ± 0.02 | **0.92 ± 0.01** | 0.87 ± 0.03 | 1.16 ± 0.02 | NA |
| Concrete | 1.23 ± 0.06 | **1.16 ± 0.02** | 0.92 ± 0.01 | **0.94 ± 0.01** | 1.00 ± 0.02 | 1.09 ± 0.01 | NA |
| Energy | 0.50 ± 0.02 | **0.47 ± 0.01** | **0.99 ± 0.00** | 0.97 ± 0.01 | 0.50 ± 0.02 | **0.47 ± 0.01** | 7% |
| Kin8nm | **1.14 ± 0.01** | 1.24 ± 0.01 | **0.97 ± 0.00** | 0.96 ± 0.00 | **1.14 ± 0.01** | 1.25 ± 0.01 | -10% |
| Naval | 0.31 ± 0.01 | **0.27 ± 0.01** | **0.99 ± 0.00** | 0.98 ± 0.00 | 0.31 ± 0.01 | **0.28 ± 0.01** | 10% |
| Power plant | 0.91 ± 0.00 | **0.86 ± 0.00** | **0.96 ± 0.00** | 0.95 ± 0.00 | 0.91 ± 0.00 | **0.86 ± 0.00** | 6% |
| Protein | 2.70 ± 0.01 | **2.28 ± 0.01** | **0.96 ± 0.00** | 0.95 ± 0.00 | 2.69 ± 0.01 | **2.27 ± 0.01** | 15% |
| Wine | 4.13 ± 0.31 | **3.13 ± 0.19** | 0.90 ± 0.01 | **0.92 ± 0.01** | 2.50 ± 0.02 | **2.33 ± 0.02** | 7% |
| Yacht | 0.31 ± 0.02 | **0.23 ± 0.02** | **0.98 ± 0.01** | 0.96 ± 0.01 | 0.30 ± 0.02 | **0.17 ± 0.00** | 43% |
| Song year | 2.90 ± NA | **2.47 ± NA** | **0.96 ± NA** | **0.96 ± NA** | 2.91 ± NA | **2.48 ± NA** | 15% |

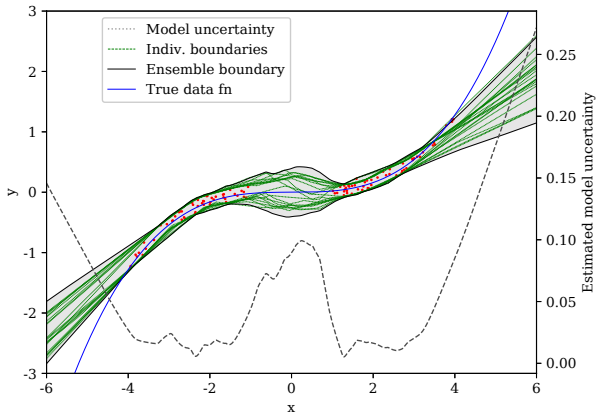

*Figure 5.* Estimation of model uncertainty with a QD ensemble.

### 6.1. Discussion

Full results of PI quality metrics are given in table 1, NLL & RMSE results are included in the appendix. Given that $Loss_{QD-soft}$ is representative of PI quality, QD-Ens outperformed MVE-Ens on all but one dataset. $PICP$ was generally closer to the 95% target, and $MPIW$ was on average 11.6% narrower. The exception to this was the *Kin8nm* dataset. In fact, this dataset was synthetic (Danafar et al., 2010), and we suspect that Gaussian noise may have been used in its simulation, which would explain the superior performance of MVE-Ens.

One drawback of QD-Ens was the fragility of the training process. Compared to MVE-Ens it required a lower learning rate, was more sensitive to decay rate, and hence needed from two to ten times more training epochs.

Other comments are as follows. We found $\lambda$ a convenient lever providing some control over $PICP$. Bootstrap resampling gave worse performance than parameter resampling, which agrees with work discussed in 4 - we suspect it would work give a much larger ensemble size. We tried to establish a relationship between the normality of residual errors and improvement of QD-Ens over MVE-Ens, but due to the variable power of normality tests analysis was unreliable.

## 7. Conclusions and Future Work

In this paper we derived a loss function for the output of PIs based on the assumption that high-quality PIs should be as narrow as possible subject to a given coverage proportion. We contrasted it with a previous work, justifying differences and showed that it can be used successfully with GD with only slight modification. We described why a single NN using the derived loss function underestimates uncertainty, and that this can be addressed by using the model in an ensemble. On ten benchmark regression datasets, the new model reduced PI widths by over 10%.

Several areas are worth further investigation: Why parameter resampling provides better performance than bootstrap resampling, how model uncertainty could be estimated through dropout or conformal prediction rather than ensembling, and the role that NN architecture plays in estimates of model uncertainty.

## Acknowledgements

The authors thank EPSRC for funding (EP/N509620/1), the Alan Turing Institute for accommodating the lead author during his work (TU/D/000016), and Microsoft for Azure credits. Personal thanks to Mr Ayman Boustati, Mr Henry-Louis de Kergorlay, and Mr Nicolas Anastassacos.

# A. Experimental details

In this section we give full experimental details of the work described in the main paper. Code is made available online[7]. Note that whilst single-layer NNs were used to be consistent with previous works, the developed methods may be applied without modification to deeper architectures.

## A.1. Qualitative Experiments

### A.1.1. TRAINING METHOD: PSO VS. GD

For the qualitative training method comparison, PSO vs. GD (section 5.1), NNs used ReLU activations and 50 nodes in one hidden layer. GD was trained using $Loss_{QD-soft}$ and run for 2,000 epochs, PSO was trained using $Loss_{QD}$ and run for 50 particles over 2,000 iterations, parameters as given in SPSO 2011 were followed (Thomas et al., 2012). Data consisted of 200 points sampled uniformly from the interval $[-2, 2]$.

### A.1.2. LOSS FUNCTION: QD VS. MVE

For the loss function comparison, QD vs. MVE (section 5.2), NNs used Tanh activations and 50 nodes in one hidden layer. Both methods were trained with GD and results are for an individual NN (not ensembled). Data consisted of 200 points sampled uniformly from the interval $[-2, 2]$.

### A.1.3. MODEL UNCERTAINTY ESTIMATION: ENSEMBLES

For evaluation of ensembling (section 5.3), we sampled 50 points uniformly in the interval $[-4, -1]$, and another 50 from $[1, 4]$. An ensemble of ten QD NNs using ReLU activations and 50 nodes in one hidden layer was trained with GD, using parameter resampling.

### A.1.4. TRAINING METHOD / LOSS FUNCTION PERMUTATION

We provide quantitative results in table 2 for the two synthetic datasets described in sections 5.1 & 5.2. These results cover permutations of loss function and training method [LUBE, QD, MVE]x[GD, PSO], using individual NNs (not ensembled).

Again, 200 training data points were sampled uniformly from the interval $[-2, 2]$. Validation was on 2,000 data points sampled from the same interval. Experiments were repeated ten times. PIs targeted 95% coverage.

LUBE relates to the 'softened' version of eq. (16), and QD to $Loss_{QD-soft}$. Softened versions were used for both GD and PSO (note 'soft' and 'hard' versions were contrasted in section 5.1).

---

[7] https://github.com/TeaPearce

For GD, 2,000 epochs were used, for PSO, 10 particles were run over 2,000 epochs. This meant the computational effort for PSO was five times that required by GD - the computational equivalent of 2,000 forward and backward passes for GD is 2 particles at 2,000 epochs for PSO.

NLL & RMSE metrics were computed, however should be viewed with caution for LUBE and QD - see section A.2.2.

All training methods and loss functions slightly overfitted the training data, producing PICP's lower than 95%. Generally GD-trained NNs outperformed their PSO counterparts in terms of the primary metric of their loss function ($MPIW$ for LUBE and QD, NLL for MVE), although quality of other metrics was similar. QD produced comparable results to LUBE - we note that our contributions to the loss function were from a theoretical and usability perspective and did not expect a large impact on performance. MVE produced PIs of comparable width to LUBE and QD for the case of normal noise, but PIs were significantly wider in the exponential noise case.

## A.2. Benchmarking Experiments

### A.2.1. SET UP AND HYPERPARAMETERS

For the benchmarking section, experiments were run across ten open-access datasets, train/test folds were randomly split 90%/10%, with experiments repeated 20 times, input and target variables were normalised to zero mean and unit variance. NNs had 50 neurons in one hidden layer with ReLU activations. The exceptions to this were for experiments with the two largest datasets, *Protein* and *Song Year*, where NNs had 100 neurons in one hidden layer, and were repeated five times and one time respectively.

The softening factor was constant for all datasets, $s = 160.0$. For the majority of the datasets $\lambda = 15.0$, but was set to $4.0$ for *naval*, $40.0$ for *protein*, $30.0$ for *wine*, and $6.0$ for *yacht*. The Adam optimiser was used with batch sizes of 100. Five NNs were used in each ensemble, using parameter resampling.

Hyperparameters requiring tuning were learning rate, decay rate, $\lambda$, initialising variance, and number of training epochs. Tuning was done on a single 80%/20% train/validation split and using random search.

### A.2.2. NLL & RMSE RESULTS

In table 3 we report NLL & RMSE in unnormalised form to be consistent with previous works. Note that in the main results (section 6) we found it more meaningful to leave $MPIW$ in normalised form so that comparisons could be made across datasets.

To compute NLL & RMSE for QD-Ens, we used the midpoint of the PIs as the point estimate to calculate RMSE.

*Table 2.* Full permutation results of training method and loss function on synthetic data with differing noise distributions. Mean ± one standard error.

| | LUBE | | MVE | | QD | |
|---|---|---|---|---|---|---|
| | GD | PSO | GD | PSO | GD | PSO |
| NORMAL NOISE | | | | | | |
| PICP | 0.90 ± 0.01 | 0.91 ± 0.01 | 0.91 ± 0.01 | 0.93 ± 0.01 | 0.91 ± 0.01 | 0.91 ± 0.01 |
| MPIW | 1.16 ± 0.05 | 1.13 ± 0.04 | 1.11 ± 0.04 | 1.00 ± 0.02 | 0.97 ± 0.04 | 1.07 ± 0.04 |
| RMSE | 0.42 ± 0.01 | 0.43 ± 0.01 | 0.39 ± 0.00 | 0.38 ± 0.01 | 0.40 ± 0.01 | 0.41 ± 0.01 |
| NLL | 0.60 ± 0.07 | 0.47 ± 0.06 | -0.44 ± 0.02 | -0.23 ± 0.04 | 0.13 ± 0.08 | 0.37 ± 0.08 |
| EXPONENTIAL NOISE | | | | | | |
| PICP | 0.91 ± 0.01 | 0.92 ± 0.01 | 0.93 ± 0.01 | 0.95 ± 0.00 | 0.91 ± 0.01 | 0.93 ± 0.01 |
| MPIW | 0.80 ± 0.02 | 0.93 ± 0.04 | 1.07 ± 0.04 | 0.99 ± 0.03 | 0.84 ± 0.03 | 0.98 ± 0.04 |
| RMSE | 0.39 ± 0.00 | 0.40 ± 0.01 | 0.38 ± 0.01 | 0.37 ± 0.01 | 0.39 ± 0.01 | 0.41 ± 0.01 |
| NLL | 0.36 ± 0.09 | 0.64 ± 0.13 | -0.48 ± 0.02 | -0.18 ± 0.02 | 0.40 ± 0.05 | 0.54 ± 0.13 |

We computed the equivalent Gaussian distribution of the PIs by centering around this midpoint and using a standard deviation of $(y_{Ui} - y_{Li})/3.92$ (since the PI represented 95% coverage), which enabled NLL to be computed. We emphasise that by doing this, we break the distribution-free assumption of the PIs, and include these purely for the purpose of consistency with previous work. Unsurprisingly, NLL & RMSE metrics for QD-Ens are poor. MVE-Ens results are in line with previously reported work (Lakshminarayanan et al., 2017).

*Table 3.* RMSE and NLL on ten benchmarking regression datasets; mean $\pm$ one standard error, best result in bold.

|  |  |  | RMSE | | NLL | |
|---|---|---|---|---|---|---|
|  | $n$ | $D$ | MVE-ENS | QD-ENS | MVE-ENS | QD-ENS |
| BOSTON | 506 | 13 | **2.84 $\pm$ 0.19** | 3.38 $\pm$ 0.26 | **2.60 $\pm$ 0.10** | 2.74 $\pm$ 0.14 |
| CONCRETE | 1,030 | 8 | **5.20 $\pm$ 0.10** | 5.76 $\pm$ 0.10 | **2.95 $\pm$ 0.04** | 3.10 $\pm$ 0.02 |
| ENERGY | 768 | 8 | **1.67 $\pm$ 0.05** | 2.30 $\pm$ 0.04 | **1.12 $\pm$ 0.05** | 1.62 $\pm$ 0.06 |
| KIN8NM | 8,192 | 8 | **0.08 $\pm$ 0.00** | 0.09 $\pm$ 0.00 | **-1.28 $\pm$ 0.01** | -1.14 $\pm$ 0.01 |
| NAVAL | 11,934 | 16 | **0.00 $\pm$ 0.00** | **0.00 $\pm$ 0.00** | **-5.67 $\pm$ 0.03** | -5.73 $\pm$ 0.03 |
| POWER PLANT | 9,568 | 4 | **3.94 $\pm$ 0.03** | 4.10 $\pm$ 0.03 | **2.77 $\pm$ 0.01** | 2.83 $\pm$ 0.01 |
| PROTEIN | 45,730 | 9 | **4.35 $\pm$ 0.02** | 4.98 $\pm$ 0.02 | **2.74 $\pm$ 0.02** | 3.12 $\pm$ 0.02 |
| WINE | 1,599 | 11 | **0.62 $\pm$ 0.01** | 0.65 $\pm$ 0.01 | **1.07 $\pm$ 0.06** | 1.15 $\pm$ 0.03 |
| YACHT | 308 | 6 | 1.36 $\pm$ 0.09 | **1.00 $\pm$ 0.08** | 1.02 $\pm$ 0.05 | **0.76 $\pm$ 0.07** |
| SONG YEAR | 515,345 | 90 | **8.88 $\pm$ NA** | 9.30 $\pm$ NA | **3.37 $\pm$ NA** | 3.58 $\pm$ NA |